# Semantic Segmentation and Data Fusion of Microsoft Bing 3D Cities and Small UAV-based Photogrammetric Data

**Meida Chen, Andrew Feng, Kyle McCullough, Pratusha Bhuvana Prasad,**
*USC Institute for Creative Technologies*
**Los Angeles, California**
{mechen, feng, McCullough, bprasad}@ict.usc.edu

**Ryan McAlinden**
*Synthetic Training Environment Cross Functional Team*
**Los Angeles, California**
ryan.e.mcalinden.civ@mail.mil

**Lucio Soibelman**
*USC Department of Civil and Environmental Engineering*
**Los Angeles, California**
soibelman@usc.edu

## ABSTRACT

With state-of-the-art sensing and photogrammetric techniques, Microsoft Bing Maps team has created over 125 highly detailed 3D cities from 11 different countries that cover hundreds of thousands of square kilometer areas. The 3D city models were created using the photogrammetric technique with high-resolution images that were captured from aircraft-mounted cameras. Such a large 3D city database has caught the attention of the US Army for creating virtual simulation environments to support military operations. However, the 3D city models do not have semantic information such as buildings, vegetation, and ground and cannot allow sophisticated user-level and system-level interaction. At I/ITSEC 2019, the authors presented a fully automated data segmentation and object information extraction framework for creating simulation terrain using UAV-based photogrammetric data (Chen et al. 2019). This paper discusses the next steps in extending our designed data segmentation framework for segmenting 3D city data. In this study, the authors first investigated the strengths and limitations of the existing framework when applied to the Bing data. The main differences between UAV-based and aircraft-based photogrammetric data are highlighted. The data quality issues in the aircraft-based photogrammetric data, which can negatively affect the segmentation performance, are identified. Based on the findings, a workflow was designed specifically for segmenting Bing data while considering its characteristics. In addition, since the ultimate goal is to combine the use of both small unmanned aerial vehicle (UAV) collected data and the Bing data in a virtual simulation environment, data from these two sources needed to be aligned and registered together. To this end, the authors also proposed a data registration workflow that utilized the traditional iterative closest point (ICP) with the extracted semantic information.

## ABOUT THE AUTHORS

**Meida Chen** is currently a research associate at the University of Southern California's Institute for Creative Technologies (USC-ICT) working on the One World Terrain project. He received his Ph.D. degree at USC Sonny Astani Department of Civil and Environmental Engineering. His research focuses on the semantic modeling of outdoor scenes for the creation of virtual environments and simulations. Email: mechen@ict.usc.edu

**Andrew Feng** is currently a research scientist at USC-ICT working on the One World Terrain project. Previously, he was a research associate focusing on character animation and automatic 3D avatar generation. His research work involves applying machine learning techniques to solve computer graphics problems such as animation synthesis, mesh skinning, and mesh deformation. He received his Ph.D. and MS degree in computer science from the University of Illinois at Urbana-Champaign. Email: feng@ict.usc.edu

**Kyle McCullough** is the Director of Modeling & Simulation at USC ICT. His research involves geospatial initiatives in support of the Army's One World Terrain project, as well as advanced prototype systems development. His work includes utilizing AI and 3D visualization to increase fidelity and realism in large-scale dynamic simulation environments, and automating typically human-in-the-loop processes for Geo-specific 3D terrain data generation. He has published multiple papers on photogrammetric reconstruction, automated feature attribution, and autonomous agents. Kyle received awards from I/ITSEC and the Raindance festival, winning "Best Interactive Narrative VR Experience" in 2018. He has a BFA from New York University.
Email: McCullough@ict.usc.edu

**Pratusha Bhuvana Prasad** is currently a researcher at USC-ICT working on the One World Terrain project. Her research focuses on computer vision for geometry and using machine learning methods to solve the same. She has a Master's degree from the Ming Hsieh Department of Electrical and Computer Engineering, USC. Email: bprasad@ict.usc.edu

**Ryan McAlinden** is a Senior Technology Advisor for the US Army's Synthetic Training Environment (STE), part of Army Futures Command (AFC). He is the Cross Functional Team (CFT) lead for the One World Terrain (OWT) initiative, which seeks to produce a high-resolution, geo-specific 3D representation of the surface used in the latest rendering engines, simulations and applications. Ryan previously served as Director of Modeling, Simulation & Training at the University of Southern California's Institute for Creative Technologies where he led several initiatives related to training modernization across the Services. Ryan rejoined ICT in 2013 after an assignment at the NATO Communications & Information Agency (NCIA) in The Hague, Netherlands. There he led the provision of operational analysis support to the International Security Assistance Force (ISAF) Headquarters in Kabul, Afghanistan. Ryan's research interests lie in the design, development, implementation and fielding of solutions that are at the cross-section of 3D rendering, geospatial science, and human-computer interaction. Ryan earned his B.S. from Rutgers University and M.S. in computer science from USC. Email: ryan.e.mcalinden.civ@mail.mil

**Lucio Soibelman** is a Professor and Chair of the Sonny Astani Department of Civil and Environmental Engineering at USC. Dr. Soibelman's research focuses on use of information technology for economic development, information technology support for construction management, process integration during the development of large-scale engineering systems, information logistics, artificial intelligence, data mining, knowledge discovery, image reasoning, text mining, machine learning, multi-reasoning mechanisms, sensors, sensor networks, and advanced infrastructure systems. Email: soibelman@usc.edu

# Semantic Segmentation and Data Fusion of Microsoft Bing 3D Cities and Small UAV-based Photogrammetric Data

| | | |
|---|---|---|
| **Meida Chen, Andrew Feng, Kyle McCullough, Pratusha Bhuvana Prasad,** *USC Institute for Creative Technologies* Los Angeles, California {mechen, feng, McCullough, bprasad}@ict.usc.edu | **Ryan McAlinden** *Synthetic Training Environment Cross Functional Team* Los Angeles, California ryan.e.mcalinden.civ@mail.mil | **Lucio Soibelman** *USC Department of Civil and Environmental Engineering* Los Angeles, California soibelman@usc.edu |

## INTRODUCTION

The Microsoft Bing Maps team has devoted considerable efforts to generating 3D city models from over 125 different cities in 11 different countries. In addition, the team is continuously expanding its 3D terrain maps to incorporate more cities and countries, along with improving the maps' photorealism by compositing 2D textures from aerial photography. Chen et al. (Chen et al., Forthcoming; Chen et al., Forthcoming; Chen et al., 2019) showcased an automatic segmentation pipeline on 3D photogrammetry data acquired with a drone. Even with coverage of an area of 10x10 km² over 20 datasets, the authors presented superior segmentation of the terrain into the ground, vegetation, and man-made structures. However, the technique is limited to the UAV-based photogrammetric point cloud datasets. In this paper, the authors propose to widen the scope of the fully automated segmentation pipeline to utilize the coverage of the 3D data from Microsoft Bing, thereby deriving useful semantic information about the top-level terrain elements around the globe.

Creating 3D imagery of the globe requires massive efforts from 3D modeling teams, photogrammetrists, high-resolution aerial imagery, and a pipeline integrating all of the above. Another aspect to take into consideration is that teams at Microsoft Bing have already put in extensive work to merge, compress, and store zettabytes of data. Leveraging this gives us an advantage to expand segmentation and classification techniques of (Chen et al., Forthcoming; Chen et al., 2019). Technique expansion allows the recovery of semantic information (e.g., whether the 3D model is vegetation, ground or a building) on a global scale where Microsoft Bing has 3D terrain data available. Our key objectives include simulating the 3D data for rendering in a virtual environment and, hence, also includes fusing the two datasets. This fusion ensures the datasets overlap and that the regions of overlap have 3D data from only one source, which is essential.

The challenge, however, is not straightforward since the two datasets, though similar in kind, differ in modes, angles, altitude, and time of captures. storage data formats, and origins of reference. These contribute to high variance between the 3D models or point clouds generated as a result of the photogrammetry process. The deep-learning models that had already trained and tested on UAV data did not function as a plug-and-play component with the Bing data. The model needed some exposure to the Bing data during the training stage so that it could learn and reflect the attributes of the Bing 3D terrain composed of trees, man-made structures, and ground.

Our proposed approach was to expand the hierarchical deep-learning model to include large variance and diversity in the weights of the network, so that it can perform well on a global scale. We kept the segmentation pipeline automated, but designed a semi-automated rule-based approach for annotating the training dataset using Bing data to label point clouds of buildings and vegetation. A predecessor to this step would be to segment ground versus non-ground points, though this step remains the same as the variance in ground points in the two datasets is quite low, owing to 3D geometric features that are extracted through the deep-learning process. The presented study in this paper is a continued effort as part of the One World Terrain (OWT) project. For more information about the OWT project, readers can refer to http://ict.usc.edu/prototypes/one-world-terrain-owt/.

## DATA SEGMENTATION

Data segmentation and feature attribution are the critical steps for generating a realistic virtual simulation environment. Our previous work has successfully demonstrated the capability of using deep-learning approaches for segmenting 3D data and created a research prototype named Semantic Terrain Points Labeling System Plus (STPLS+). Unlike many other 3D point cloud segmentation approaches that were designed for and validated with Light Detection and Ranging (LIDAR) data, STPLS+ was specifically designed to operate on the 3D point clouds that are generated using the photogrammetric technique with low altitude UAV captures (e.g., 30 meters to 200 meters). We first applied data preprocessing steps, which included a data cleaning process and an area of interest (AOI) selection process, to remove the data noise and unwanted data. Following that, a voxelization process was applied on the cleaned point cloud to transform the unordered point set into a structured 3D voxel grid, which was then easily fed into a deep-learning architecture. STPLS+ extended an existing 2D deep-learning model architecture (e.g., U-net) to 3D by adding an additional channel to the input layer for the semantic segmentation task.

Unlike other 3D data segmentation approaches where one segmentation model is trained to predict all class labels, STPLS+ ensembles two segmentation models in a hierarchical manner. The first segmentation model is responsible for segmenting ground points and non-ground points, and the second model is responsible for segmenting the non-ground points into man-made structures and vegetation. This design decision was made considering the end use of the data for creating virtual environments.

Small errors in segmentation that may not significantly reflect in the quantitative evaluation using the F1 score and IOU can generate considerable noise that will produce a less visually pleasing 3D terrain model for simulations. Thus, by ensembling two segmentation models hierarchically, we were able to introduce carefully designed post-processing approaches into the STPLS+ framework in order to remove the small-miss segmentations and improve the visual quality. We also introduced three post-processing approaches—a ground point-cloud post-processing approach, a result-smoothing approach, and a building point-cloud-cleaning approach—within the STPLS+ framework. The ground post-processing approach utilized a connected component algorithm with the assumption that ground points should not be floating in the air, and they are close to each other to form large connected components. The result-smoothing approach utilized the well-established Conditional Random Field (CRF) algorithm to refine the labels in the segmented building and vegetation point clouds. The building point-cloud-cleaning approach used a local CRF to collect the points that were around the segmented building points with similar colors to ensure consistent appearances. Note that the building point-cloud-cleaning approach can only be used when the initial segmentation results do not have large sections with missings segmentation on vegetation (e.g., large portions of the forest are mislabeled as buildings). Please refer to (Chen et al., Forthcoming; Chen et al., 2019) for more details about STPLS+.

**Issues with Existing Training Data from Small UAV Collects**

Since STPLS+ was initially designed for photogrammetric generated data where the source images were captured with low-altitude UAV, directly applying the point cloud segmentation model—that was trained using the low-altitude UAV data to segment high-altitude aircraft data (e.g., Bing data)—may produced low accuracy results. In order to understand the strength and limitation of STPLS+ in segmenting Bing data, the authors started this research by conducting an experiment to test STPLS+ on selected Bing data.

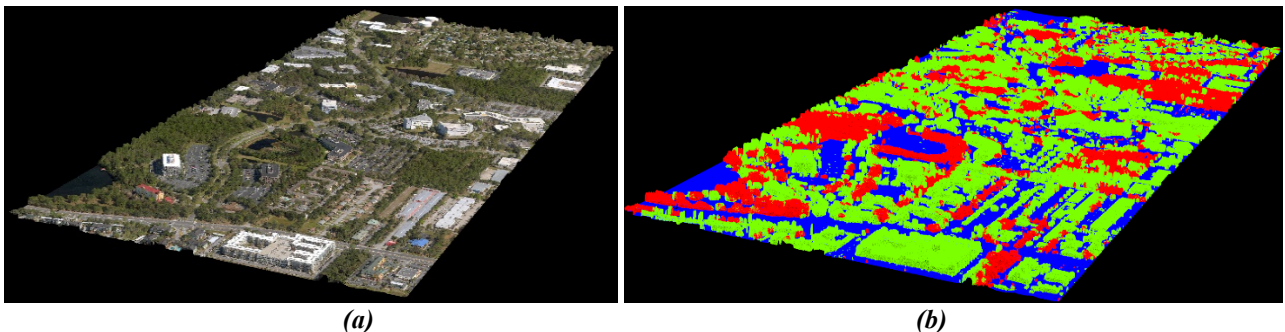

*(a)*        *(b)*

*Figure 1. Selected testing data from Bing. (a) Raw point cloud of STTC. (b) Segmentation results using STPLS+.*

A small area located in Orlando, Florida—area surrounding the Simulation and Training Technology Center (STTC)—was selected as the testing case. ***Figure 1 (a)*** shows the STTC point cloud obtained from Bing. The data covered approximately 1.7 square kilometers of area (11,993,655 points) with high- and low-rise buildings, individual trees, forests, and parks. Directly applying STPLS+ yielded low accuracy results, as expected. The segmentation results are shown in ***Figure 1 (b)***, where the ground, buildings, and trees are marked with blue, green, and red, respectively. The commonly used precision, recall, harmonic mean of precision and recall (e.g., F1 score), and intersection over union (IOU, also known as the Jaccard Index) were used to evaluate the segmentation results. IOU is computed as the *points* *overlapped* between the predicted segmentation and the ground truth divided by the *points union* between the predicted segmentation and the ground truth. ***Table 1*** summarizes the segmentation results.

**Table 1. STTC Segmentation results using STPLS+.**

| | | STTC | | | |
|---|---|---|---|---|---|
| | precision | recall | f1-score | IOU | # points |
| ground | 0.932 | 0.896 | 0.914 | 0.841 | 4,161,782 |
| building | 0.296 | 0.947 | 0.451 | 0.291 | 1,593,903 |
| tree | 0.946 | 0.439 | 0.600 | 0.428 | 6,237,970 |
| macro avg | 0.725 | 0.761 | 0.655 | 0.520 | 11,993,655 |
| weighted avg | 0.855 | 0.665 | 0.689 | 0.553 | 11,993,655 |

**What Worked Well, and What Didn't?**

The results showed clearly that the model for segmenting ground vs. non-ground points worked, but the model for segmenting building and vegetation points did not. The ground segmentation achieved 0.914 F1-score, whereas the F1-scores of vegetation and building were only 0.6 and 0.451, respectively. The precision and recall for the segmented building points are 0.296 and 0.947, respectively, and the precision and recall for the segmented vegetation points are 0.946 and 0.439, respectively. The low precision of building points and low recall of vegetation points indicated that a large number of vegetation points were mis-segmented as buildings, which is represented in *Figure 1 (b)*.

During the experiment, the authors also analyzed the pre- and post-processing approaches that were previously designed within the STPLS+ framework. It was unnecessary to use the previously designed data-preprocessing approaches for cleaning the point cloud since there were no data noises under the ground or on the boundary of the selected area of interest (AOI). The ground post-processing approach was used to effectively remove the mis-segmented points (e.g., points floating in the air) from the ground point set. The segmentation result-smoothing approach was also applied after performing the building and vegetation segmentation process to remove small artifacts. However, the building point-cloud-cleaning approach did not improve the segmentation results since a large number of tree-points were mis-segmented as buildings, which thereby made more mis-segmented tree points and worsened the situation.

**Issues with Existing Training Data from Small UAV Collects**

The main difference between Bing data and small UAV collects is that Bing data lacks details on building facades, and the quality of the Bing tree meshes is lower compared to the UAV data. Typical tree meshes from the Bing and UAV data are shown in *Figure 2 (a)* and *(b)*, respectively. Despite the fact that both Bing and UAV tree meshes appear as solid blobs without well-formed branches, the UAV data contains more details, especially at the bottom of the trees. The building and vegetation segmentation U-net model that was trained using the UAV data learned that trees do not have smooth vertical surfaces on the side. Instead, they have rough surfaces that are not perpendicular to the ground, as shown in *Figure 2 (b)*. However, as shown in *Figure 2 (a)*, the sides of the tree mesh from Bing appear as vertical surfaces that are very similar to the walls of a building. As a result, our previously trained U-net model was not able to differentiate the trees and buildings from the Bing data and mis-labeled the points on the side of the trees as buildings.

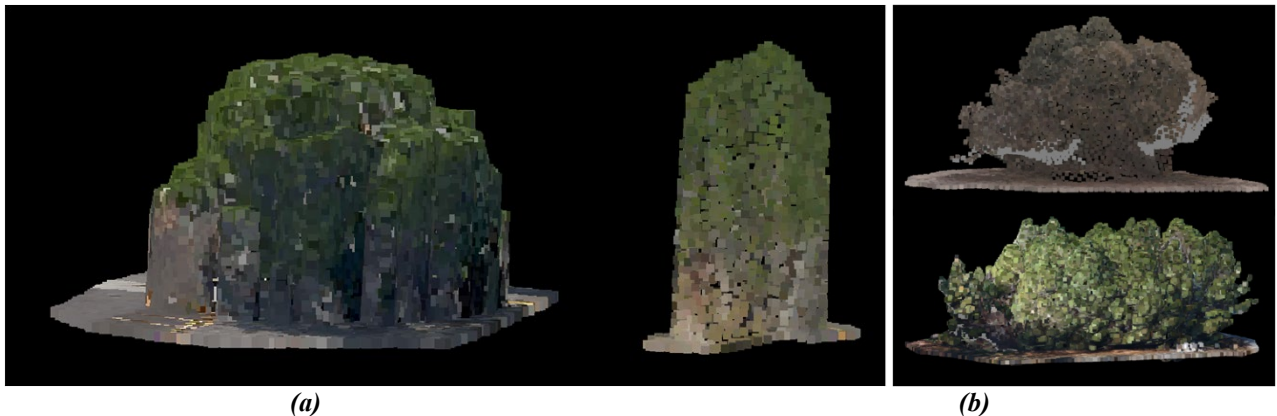

*(a)*           *(b)*

***Figure 2. Tree meshes. (a) Tree meshes from Bing. (b) Tree meshes from the UAV data.***

To overcome the challenges and improve the segmentation performance, training a U-net model that is suitable for segmenting building and vegetation points of Bing data is a necessary step. However, training a new model that can be generalized across different terrain datasets requires an extensive annotated database, and preparing such a large database is a time-consuming and labor-intensive process. To this end, a rule-based training data preparation process was designed and developed for annotating Bing data. It is worth pointing out that the authors did not intend to design a fully-automated training data preparation process. Instead, we used several rule-based approaches to facilitate and accelerate the annotation process in order to minimize users' manual labeling efforts.

**Rule-based training Data Annotation Workflow**

One strength of STPLS+ that was identified from the experiment is that the previously trained U-net model—the model that was trained using UAV data for segmenting ground and non-ground points—performed reasonably well (e.g., 0.914 F1-score) on Bing data. Thus, we utilized the ground segmentation capability of STPLS+ as the initial step to annotate the Bing data. Following that, the selected hand-crafted point features were computed and used in our designed rule-based annotation workflow. Hand-crafted point features included color features, point density features, and local-surface features that the authors previously used for point-cloud segmentation with traditional supervised machine learning algorithms (e.g., Random Forest, SVM, etc.) (Chen, McAlinden, Spicer, & Soibelman, 2019; Chen, Feng, McAlinden, & Soibelman, 2020). For color features, we simply utilized blue value from the point cloud. The density features of each point were computed as the number of points n in a sphere with a predefined radius r centered at the point. Eigenvalues are also derived for each point from the covariance matrix of its neighbor points. Eigenvalues were used to compute the local-surface feature (e.g., verticality).

$$\sum p = \frac{1}{n}\sum_{i=1}^{n}(p^i - \bar{p})(p^i - \bar{p})^T, \qquad (1)$$

Where p is a point data in the point cloud that is represented using its x, y, and z coordinates; pi is p's surrounding point; p- is the mean/center of its surrounding points. The principal component analysis was then performed on the covariance matrix to compute the eigenvalues $\lambda 1 > \lambda 2 > \lambda 3$. Please note that eigenvalues needed to be normalized between 0 to 1 with respect to $\lambda 1$. Verticality was then computed as follows:

$$\text{Verticality} = 1 - |\langle [0,0,1], e_3 \rangle|, \qquad (2)$$

Roughness is another geometric feature that was used in this research. The roughness value was computed as the distance between each point and the best fitting plane computed on its neighbor points.

With the point features discussed above, a set of designed rule-based approaches were applied on the non-ground points. It is worth pointing out here that the primary goal of using the designed rule-based approach was not to segment building and vegetation points directly but identify roof points. The identified roof points were then used to generate rough building boundaries, which were then used to extract the points within them (e.g., building points). The building points could not be annotated directly using the rule-based approach because of the tree meshes' quality issues that were identified from the previous section (e.g., the sides of the trees as vertical surfaces). Hand-crafted point features were not robust enough to differentiate the building walls and sides of trees.

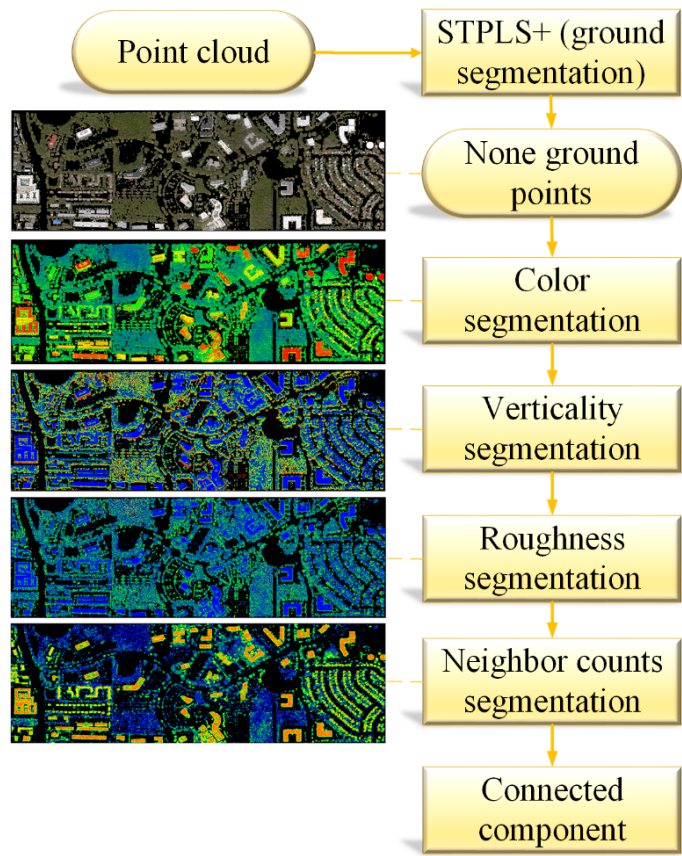

*Figure 3. Rule based roof points segmentation workflow*

The designed rule-based training data annotation workflow is shown in ***Figure 3*** using the STTC dataset as an example for illustration purposes. The workflow started with the color segmentation using the blue value, and any points with a blue value less than 60 were removed. This step partially removed tree points. Following that, verticality and roughness were computed for each point based on the remaining points. Verticality value was normalized between 0 to 1 where 1 indicated the point was located on a surface perpendicular (at a 90-degree angle) to the ground, and 0 indicated the point is located on a horizontal plane. Roughness was also normalized between 0 to 1 where 0 indicated the point is located on a flat surface, and 1 indicated the point was located on a rough surface (e.g., top of a tree). Points with verticality values above 0.5 were removed since these points are on vertical surfaces such as walls and sides of trees. Similarly, Points with roughness value above 0.3 were removed, since these points are on rough surfaces such as tops of trees.

After performing the color, verticality, and roughness segmentation processes, most of the points on trees and walls have been removed at this stage. Consequently, the point density became low in these areas. Thus, in the next step, the neighbor count of each point within the predefined radius was used. Points with less than 60 neighbor counts within 3 meters were removed. The remaining points include most of the roof points and some noise. The last step in the designed workflow was to use a connected component algorithm to cluster the remaining points into different connected components. Noise points usually form small connected components. Thus, the connected components that had less than 100 points were removed.

The extracted roof points are shown in ***Figure 4 (a)***. Once the roof points were extracted, the next step was to segment building points based on the roofs. Intuitively, points with z values that were less than the roof height and within each roof boundary belonged to buildings. One way of achieving this was to extract the roof boundary as polygons first and then crop the points within each polygon. However, the proposed rule-based approaches could not guarantee that all points on the roofs could be segmented (e.g., missing points on the edge of a roof). Thus, in practice, the authors utilized the k-nearest neighbor algorithm in 2D space. First, for each roof point, we extracted its neighbor points within 3 meters from all other points using only x and y coordinates. Following that, for each of the extracted points, we labeled it as a building point only if its z value was smaller than its roof neighbor point. The reason we used z value to remove points above the roof is that for low-rise buildings, it is possible to have points on tree branches/leaves hanging over the roofs. The extracted building points are shown in ***Figure 4 (b)***.

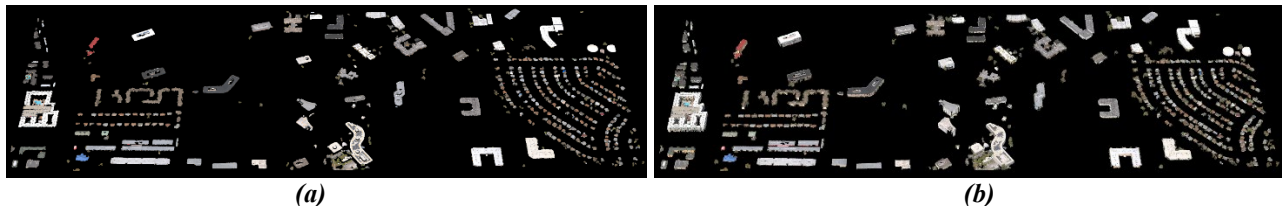

*(a)*                                              *(b)*

*Figure 4. Building segmentation using roof points. (a) the extracted roof points. (b) the segmented building points.*

***Figure 5*** shows the result of performing the designed rule-based data annotation approach on the STTC dataset. As mentioned before, the rule-based data annotation workflow was designed to minimize users' manual labeling efforts but cannot be used to generate the ground truth data automatically. Therefore, users have to make corrections on the mis-segmentation parts manually to create the ground truth.

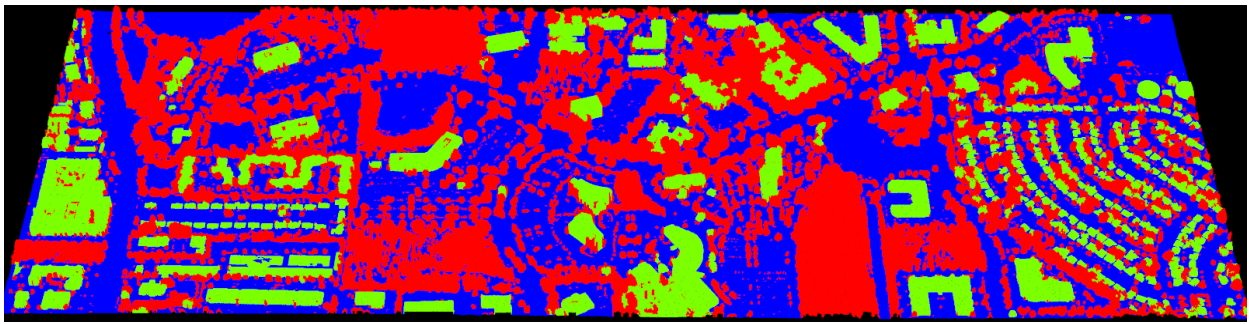

*Figure 5. Annotated STTC dataset using the proposed rule-based data annotation workflow.*

In order to quantify the manual efforts saved by using our designed workflow, a quantitative analysis was conducted to compare the annotated STTC dataset using the proposed rule-based data annotation workflow against the ground truth that was created after the user corrections. ***Table 2*** summarizes the comparison results. Since the designed workflow utilized STPLS+ for segmenting ground points, the segmentation result in ***Table 2*** is the same as the result that was shown in ***Table 1***. The f1-scores for building and tree segmentation were 0.844 and 0.929, respectively. The weighted average of f1-score across all three classes was 0.913, which means a user only needed to make manual corrections for about 10% of the points. In this experiment, it took a labeler four hours to make the correction for the STTC dataset.

**Table 2. STTC annotation result using the designed workflow.**

|  | STTC | | | | |
|---|---|---|---|---|---|
|  | precision | recall | f1-score | IOU | # points |
| ground | 0.932 | 0.896 | 0.914 | 0.841 | 4,161,782 |
| building | 0.805 | 0.886 | 0.844 | 0.730 | 1,593,903 |
| tree | 0.931 | 0.928 | 0.929 | 0.868 | 6,237,970 |
| macro avg | 0.889 | 0.903 | 0.896 | 0.813 | 11,993,655 |
| weighted avg | 0.915 | 0.911 | 0.913 | 0.841 | 11,993,655 |

In order to improve the performance of STPLS+ on segmenting Bing data, we annotated two other datasets—downtown Boston and Syracuse—for creating the Bing database. It is worth pointing out that some of the threshold values needed to be changed when annotating both Boston and Syracuse datasets using the proposed rule-based data annotation workflow. For instance, when annotating the Boston dataset, we used the same threshold values as the STTC dataset for color, verticality, and roughness segmentation processes. Still, we changed the neighbor counts to 20 during the neighbor counts segmentation step. The RGB point clouds and color-coded annotations of Boston and Syracuse are shown in ***Figure 6***.

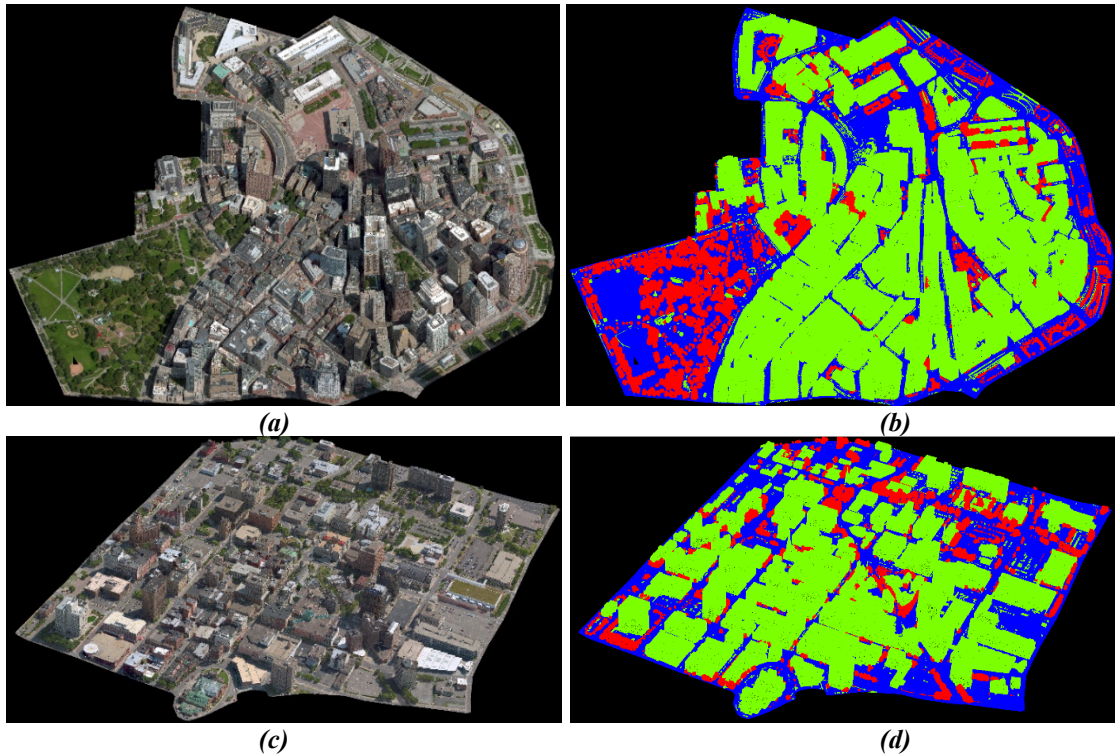

*(a)*          *(b)*

*(c)*          *(d)*

***Figure 6. Bing data annotations.***

**Data segmentation experiments and Results**

To quantify the performance improvement of STPLS+, we used the Boston and Syracuse data as the training sets and tested on the STTC dataset again. Since we did not have a huge dataset for training, instead of training a U-net model from scratch, we fine-tuned the model that was previously trained using the low-altitude UAV data. In addition, we used a simple yet effective data augmentation strategy to increase the size of the training data. Both point clouds were rotated horizontally around their center axis by an angle θ between 0 to 360 degrees to produce more training data. θ was set to 60 degrees in this study. Note that since we were using the previously trained ground segmentation model, we only fine-tuned the model for segmenting buildings and trees. The model was trained for 60 epochs with a minibatch size of 6. A widely used optimization algorithm, e.g., -Adam (Kingma and Ba 2014), was used in this study to update network weights during training. The θ value and epoch size were selected so that the model could be trained within a reasonable amount of time (e.g., less than 24 hours).

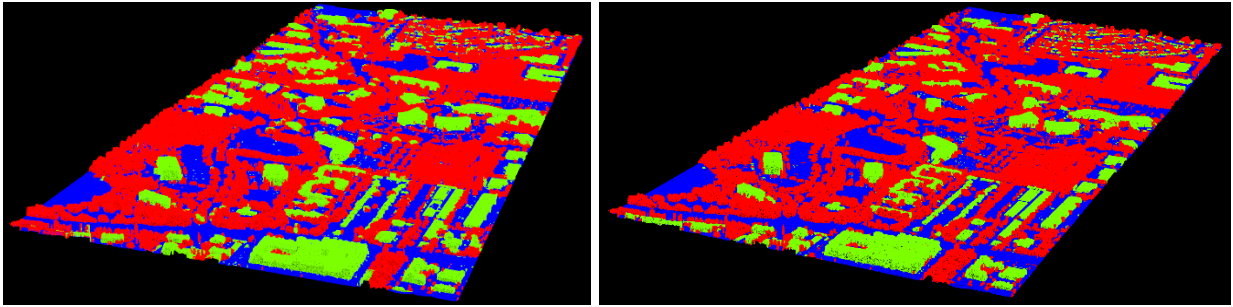

*Figure 7. STTC ground truth and segmentation result. (a) Segmentation result using the fine-tuned model. (b) Ground truth annotation.*

The segmentation results using STPLS+ with the fine-tuned model is shown in ***Figure 7 (a)***, and the ground truth STTC annotation is shown in ***Figure 7 (b)***. ***Table 3*** summarizes the quantitative analysis of the segmentation results. Compared to the result we previously showed in ***Table 1***, building and tree segmentation performance was dramatically improved using the new Bing training data (e.g., F1-score improved from 0.451 to 0.798 in the building points, from 0.60 to 0.919 for the tree points). The result indicated that the STPLS+ could work in a fully automated manner for Bing data segmentation the same way as it was originally designed for UAV data. While the results in ***Table 2*** outperformed STPLS+, the workflow, however, cannot be used to replace STPLS+ for run-time data segmentation. This is because the designed rule-based data annotation workflow heavily relies on the defined threshold value, which varies from one area to another.

**Table 3. STTC segmentation result using the fine-tuned segmentation model.**

| | precision | recall | f1-score | IOU | # points |
|---|---|---|---|---|---|
| ground | 0.932 | 0.896 | 0.914 | 0.841 | 4,161,782 |
| building | 0.718 | 0.898 | 0.798 | 0.663 | 1,593,903 |
| tree | 0.938 | 0.901 | 0.919 | 0.850 | 6,237,970 |
| macro avg | 0.862 | 0.898 | 0.877 | 0.785 | 11,993,655 |
| weighted avg | 0.906 | 0.899 | 0.901 | 0.822 | 11,993,655 |

**DATA FUSION**

Alongside segmenting Bing data for creating virtual environments, the fusion of Bing and UAV data was also a necessary step to create usable data for simulation on a large scale. Regardless of the different levels of detail and different accuracy of both datasets, the designed fusion process aims to register the two datasets into the same coordinate system to create a single seamless 3D model that serves the mission. In this study, the authors first designed a data fusion workflow utilizing the Iterative closest point (ICP) algorithm (Besl & McKay, 1992) for point cloud registration. Several experiments were conducted to identify the limitation of the workflow. Following that, we integrated the semantic information that was extracted using STPLS+ into the workflow to improve the data fusion performance.

**Workflow without Semantic Information**

The designed data fusion workflow is shown in ***Figure 8***. The data registration was performed on the point clouds that are subsampled from the meshes. Following that, the result of the transformation was applied to the mesh data. Since both Bing data and small UAV data were georeferenced, the first step was to roughly align them using their Geoinformation.

This alignment is prone to error mainly because of the low accuracy GPS in the UAV. To further improve the registration accuracy, the ICP algorithm was used. Please note that ICP was used twice to maximize its registration performance during the fine registration process. After the first time of using ICP, the overlapped portion of Bing data was segmented with a boundary buffer of 20 meters. Following that, during the second iteration of ICP, only the segmented area of Bing data was used. Once the two point clouds were aligned, the overlapped Bing meshes were removed so that only one data source existed in a location.

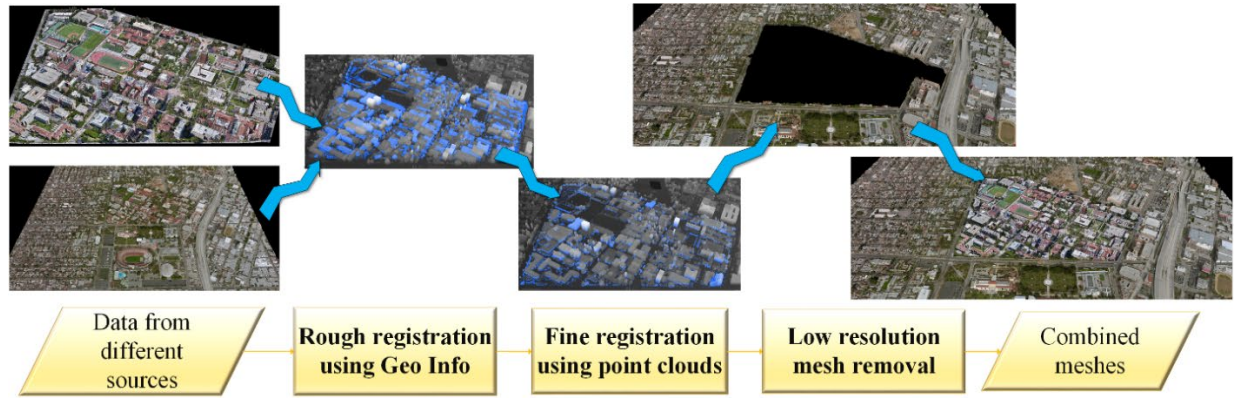

*Figure 8. Data fusion workflow.*

The designed workflow was tested with five different datasets including 1) USC Coliseum, 2) USC Galen Center, 3) Orlando Convention Center, 4) USC campus, and 5) Fort Story. The results shown in **Figure 9**. demonstrate that our designed data fusion workflow could be used to align the Bing and small UAV data to some extent. For small areas such as USC Coliseum, USC Galen Center, and Orlando Convention Center—where the drone data was not distorted and the ground in the drone data was flat and matched the Bing data—a reasonably well data alignment was achieved using the designed workflow.

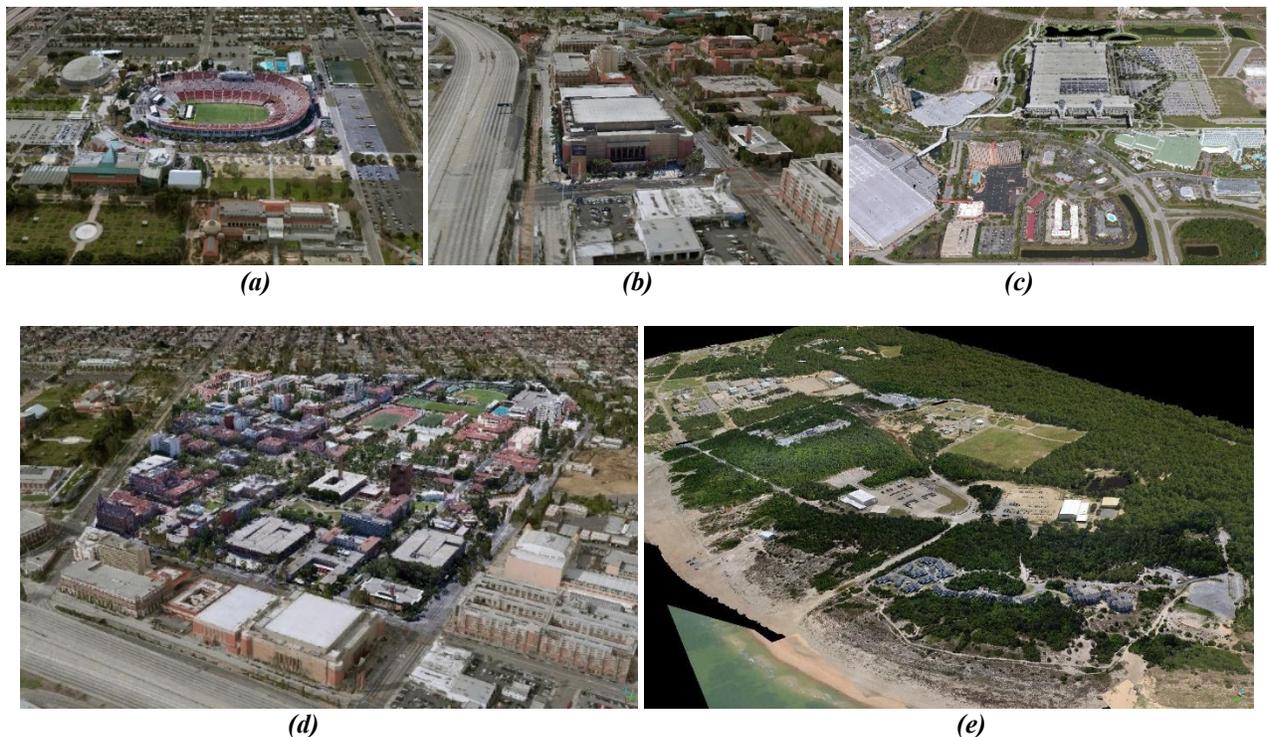

*Figure 9. Data fusion results. (a) USC Coliseum. (b) USC Galen Center. (c) Orlando Convention Center. (d) USC campus. (e) Fort Story*

However, for larger areas such as the USC campus and Fort Story—where the UAV data was distorted and the shape of the ground did not match the Bing data—there were still gaps between the two datasets, as shown in ***Figure 10 (a) and (b)***. In such cases, the UAV-captured terrain was bowl-shaped, and the Bing data was flat. Consequently, when the two datasets were registered, the center of the UAV data was below the Bing data, and the boundary of the UAV data was above the Bing data (e.g., the gaps shown in ***Figure (10)***. The misaligned data at the center did not create an issue since the Bing data was removed, and only the UAV data existed after the process. However, the misaligned data at the boundary (e.g., the gap between the two datasets) caused issues when doing ground-level simulation. As a result, players could not navigate on the ground between the two datasets because of the gap.

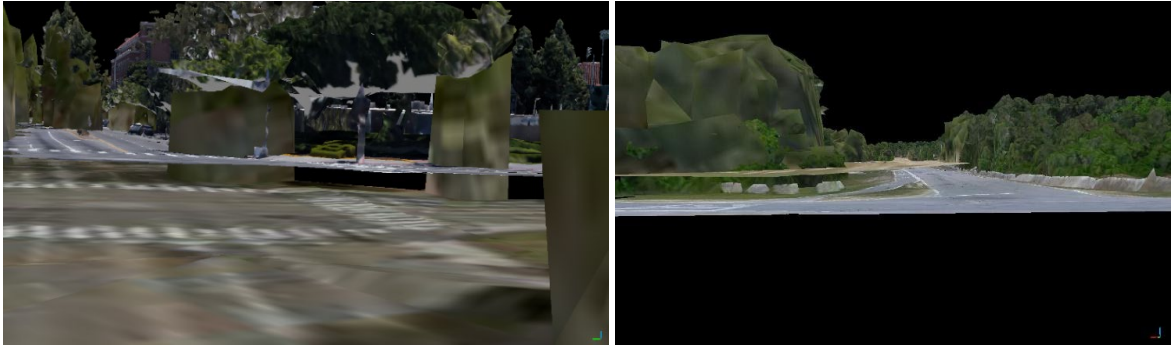

*(a)* *(b)*

***Figure 10. Data fusion issues. (a) USC campus. (b) Fort Story***

**Workflow with Semantic Information**

In order to overcome the challenge mentioned above and improve the performance of the designed data fusion workflow, we utilized the semantic information that was extracted from STPLS+ for the registration process. We also modified the fine registration process in the designed workflow. In addition to performing the ICP algorithm twice on the overlapped portions of the two datasets, we segmented the two datasets into the ground, and non-ground point sets. Following that, we reran the ICP algorithm on the segmented ground points. Since the issue we were trying to solve was located near the boundary of the data, we only used the border of the ground data for the registration process. Note that there were no vertical objects in the ground point set, and registering the two ground point sets did not produce accurate translation on the x- and y-axis in the transformation matrix. Thus, the x and y translation was set to zero in the transformation matrix. This transformation matrix was used as the final registration matrix to align the two datasets.

The authors also attempted to use STPLS+ to segment buildings for computing more accurate translation on the x- and y-axis. However, since the segmentation results contained noise that were at very different locations in the UAV and Bing data, the computed x and y translation was not robust across different areas. Thus, only ground registration was used in this study. ***Figure 11*** shows the results of using the data fusion workflow with the semantic information on the USC campus and Fort Story datasets. The size of the gap between the two datasets were decreased but were not removed entirely. This is because the ground of the UAV dataset was not as flat as Bing data, rigid point cloud registration algorithms such as ICP could not perfectly align the two datasets. Thus, future research is still needed to explore the non-rigid registration on solving the issue.

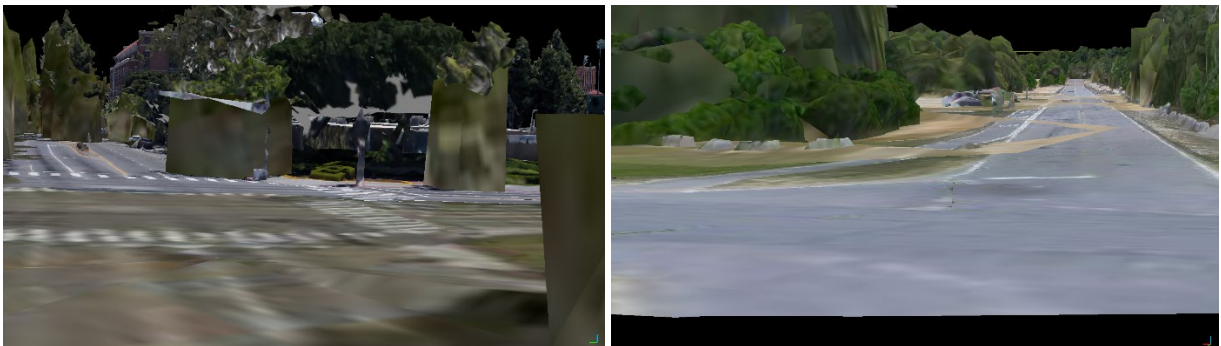

*(a)* *(b)*

***Figure 11. Data fusion results with semantic information. (a) USC campus. (b) Fort Story***

**CONCLUSION**

In this paper, we have presented a system to broaden the reach of our segmentation model on a global scale. By utilizing existing 3D terrain data of the world from Microsoft Bing and annotating Bing data from 3 different cities semi-automatically, our model exhibits robust semantic segmentation capabilities. We first showed that our model is able to perform fairly for a distinction between the ground and non-ground points over the Bing dataset without any modifications. We next presented rule-based approaches, namely segmentation based on color, verticality and roughness, followed by point-density filtering to classify data points as either vegetation or man-made structures. We showed this process is automatic and vital to achieve a weighted average f-1 score of 0.913. Once data is annotated to 90% accuracy, in order to train our model with absolute ground truth, we correct the remaining 10% and fine tune our STPLS+ model to achieve a testing f-1 score of 0.901, outperforming the original model.

In addition, we also demonstrated the applicability of our proposed method for simulations in functional virtual synthetic environments with a data fusion workflow. As part of this method, rendering the Microsoft Bing data in alignment with the UAV dataset is a powerful combination of the two datasets that can be used for large-scale virtual hands-on training. Our method also uses georeference supported registration and ICP to roughly align the point clouds and our segmentation process to facilitate alignment near the boundaries. Currently, our qualitative results present well-aligned terrain, where the region is flat and data from either dataset is not distorted. Futurework in this direction constitutes reducing the gap in alignment-overlap between datasets in challenging terrain.

**ACKNOWLEDGEMENTS**

The authors would like to thank the two primary sponsors of this research: Army Futures Command (AFC) Synthetic Training Environment (STE) and the Office of Naval Research (ONR). We would also like to acknowledge the assistance provided by 3/7 Special Forces Group (SFG), Naval Special Warfare (NSW), the National Training Center (NTC), and the U.S. Marine Corps. This work is supported by University Affiliated Research Center (UARC) award W911NF-14-D-0005. Statements and opinions expressed and content included do not necessarily reflect the position or the policy of the Government, and no official endorsement should be inferred.